\documentclass[10pt,twocolumn,letterpaper]{article}

\usepackage{iccv}
\usepackage{times}
\usepackage{epsfig}
\usepackage{graphicx}
\usepackage{amsmath}
\usepackage{amssymb}
\usepackage{multicol}
\usepackage{multirow}
\usepackage[table,xcdraw,dvipsnames]{xcolor}
\usepackage{booktabs}
\usepackage{adjustbox}
\usepackage{pifont}
\usepackage{comment}
\usepackage{color}
\usepackage[accsupp]{axessibility}  

\newcommand{\cmark}{\ding{51}}%
\newcommand{\xmark}{\ding{55}}%

\usepackage[pagebackref=true,breaklinks=true,letterpaper=true,colorlinks,bookmarks=false]{hyperref}
\hypersetup{citecolor=[RGB]{119,185,0}}
\usepackage[space, compress, sort]{cite}

\iccvfinalcopy 

\def\iccvPaperID{3919} 

\ificcvfinal\fi

\begin{document}

\title{FocalFormer3D : Focusing on Hard Instance for 3D Object Detection}

\author{Yilun Chen$^1$\thanks{Work done during an internship at NVIDIA.} \quad Zhiding Yu$^3$\thanks{The corresponding author is Zhiding Yu.} \quad Yukang Chen$^1$ \quad \\ Shiyi Lan$^3$ \quad
Anima Anandkumar$^{2,3}$ \quad Jiaya Jia$^1$ \quad Jose M. Alvarez$^3$ \\
$^1$The Chinese University of Hong Kong\quad $^2$Caltech \quad $^3$NVIDIA 
}

\maketitle
\ificcvfinal\thispagestyle{empty}\fi

\begin{abstract}
False negatives (FN) in 3D object detection, {\em e.g.}, missing predictions of pedestrians, vehicles, or other obstacles, can lead to potentially dangerous situations in autonomous driving. While being fatal, this issue is understudied in many current 3D detection methods. In this work, we propose Hard Instance Probing (HIP), a general pipeline that identifies \textit{FN} in a multi-stage manner and guides the models to focus on excavating difficult instances. For 3D object detection, we instantiate this method as FocalFormer3D, a simple yet effective detector that excels at excavating difficult objects and improving prediction recall. FocalFormer3D features a multi-stage query generation to discover hard objects and a box-level transformer decoder to efficiently distinguish objects from massive object candidates. Experimental results on the nuScenes and Waymo datasets validate the superior performance of FocalFormer3D. The advantage leads to strong performance on both detection and tracking, in both LiDAR and multi-modal settings. Notably, FocalFormer3D achieves a 70.5 mAP and 73.9 NDS on nuScenes detection benchmark, while the nuScenes tracking benchmark shows 72.1 AMOTA, both ranking 1st place on the nuScenes LiDAR leaderboard.
Our code is available at \url{https://github.com/NVlabs/FocalFormer3D}.

\end{abstract}


\begin{figure}[t!]
	\begin{center}
		\includegraphics[width=\linewidth]{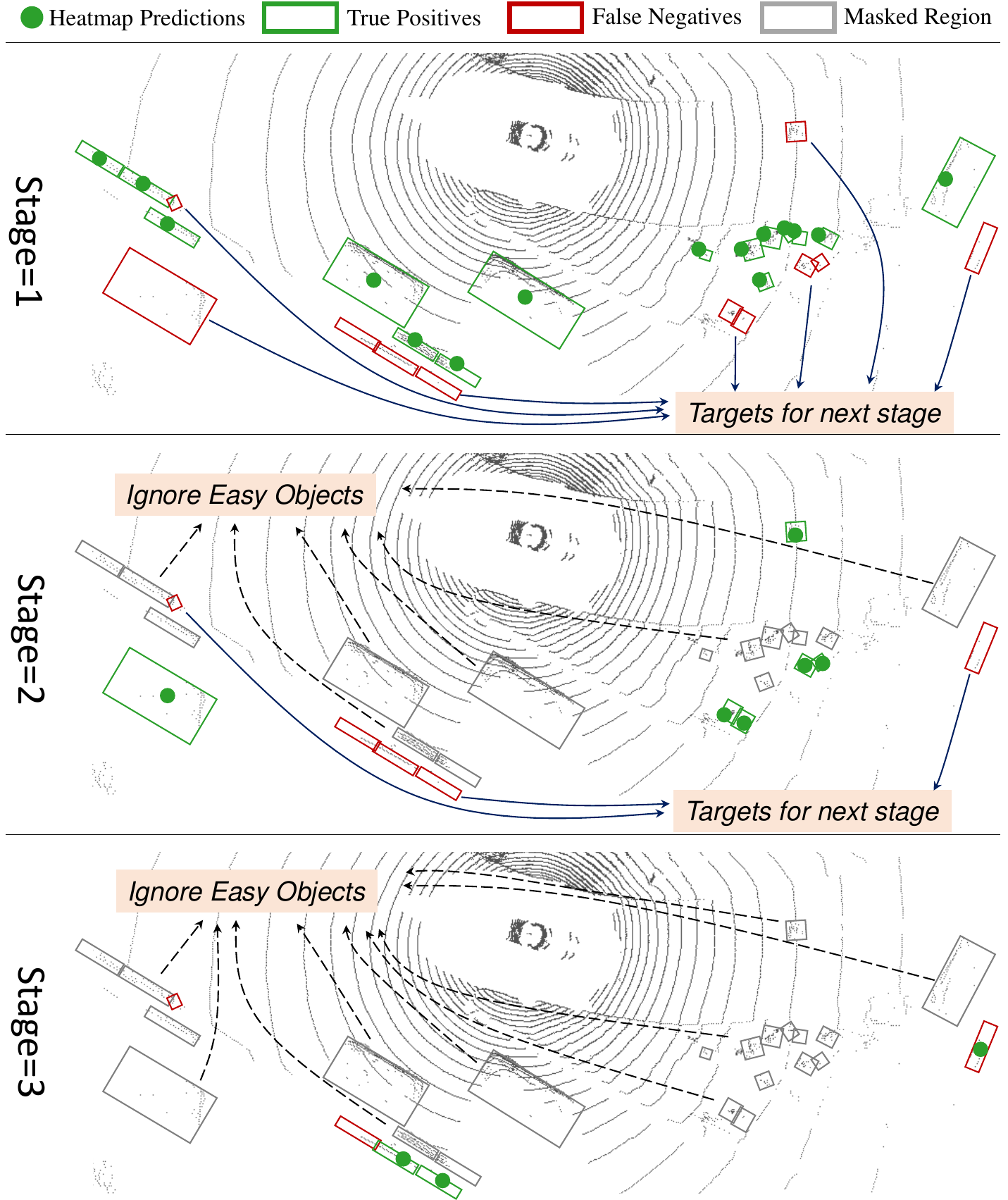}
	\end{center}
	\caption{\textbf{Visual example for Hard Instance Probing (HIP).} By utilizing this multi-stage prediction approach, our model can progressively focus on hard instances and facilitate its ability to gradually detect them. At each stage, the model generates some \textit{Positive} object candidates (represented by green circles). Object candidates assigned to the ground-truth objects can be classified as either \textit{True Positive}s (\textit{TP}, represented by  \textcolor[rgb]{0.41,0.55,0.13}{green boxes}) and \textit{False Negative}s (\textit{FN}, represented by {\color{red} red boxes}) during training. We explicitly model the unmatched ground-truth objects as the hard instances, which become the main targets for the subsequent stage. Conversely, \textit{Positive}s are considered easy samples (represented by {\color{gray} gray boxes}) and will be ignored in subsequent stages at both training and inference time. At last, all heatmap predictions across stages are collected as the initial object candidates. We ignored the \textit{False Positive}s for better visualizations.}\label{fig:visualize pipeline}\vspace{-2mm}
\end{figure}

\section{Introduction}
3D object detection is an important yet challenging perception task. Recent state-of-the-art 3D object detectors mainly rely on bird's eye view (BEV) representation~\cite{VoxelNet, pointpillar, centerpoint}, where features from multiple sensors are aggregated to construct a unified representation in the ego-vehicle coordinate space. There is a rich yet growing literature on BEV-based 3D detection, including multi-modal fusion~\cite{transfusion, bevfusion, bevfusionmit, li2022uvtr, li2022vff, jiao2022msmdfusion, deepinteraction}, second-stage refinements (surface point pooling~\cite{centerpoint}, RoIPool~\cite{pointrcnn, fastpointrcnn, voxelrcnn, lidarrcnn}, and cross attention modules~\cite{transfusion, groupfree}). 

Despite the tremendous efforts, there has been limited exploration to explicitly address {\em false negatives or missed objects} often caused by occlusions and clutter background. False negatives are particularly concerning in autonomous driving as they cause missing information in the prediction and planning stacks.
When an object or a part of an object is not detected, this can result in the autonomous vehicle being unaware of potential obstacles such as pedestrians, cyclists, or other vehicles. This is especially hazardous when the vehicle is moving at high speeds and can lead to potentially dangerous situations. Therefore, reducing false negatives is crucial to ensure the safety of autonomous driving.

To address the challenge of \textit{False Negative}s in 3D detection, we propose and formulate a pipeline called \textit{Hard Instance Probing} (HIP). Motivated by cascade-style decoder head for object detection~\cite{cascadercnn, deformabledetr, dndetr}, we propose a pipeline to probe false negative samples progressively, which significantly improves the recall rate Fig.~\ref{fig:visualize pipeline} illustrates the pipeline in a cascade manner. In each stage, HIP suppresses the true positive candidates and focuses on the false negative candidates from the previous stages. By iterating the HIP stage, our approach can save those hard false negatives.

\begin{figure*}
	\begin{center}
		\includegraphics[width=1.\textwidth]{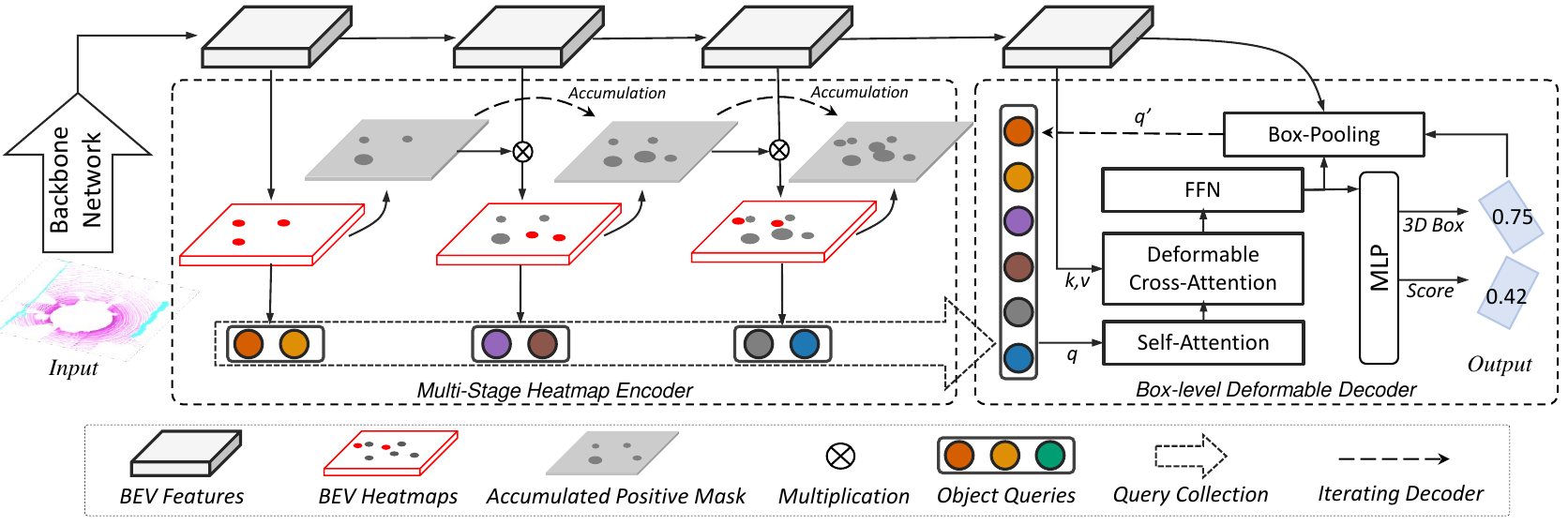}
	\end{center}
	\caption{\textbf{Overall architecture of FocalFormer3D}. The overall framework comprises two novel components: a multi-stage heatmap encoder network that uses the Hard Instance Probing (HIP) strategy to produce high-recall object queries (candidates), and a deformable transformer decoder network with rescoring mechanism that is responsible for eliminating false positives from the large set of candidates. (a) Following feature extraction from modalities, the map-view features produce a set of multi-stage BEV features and then BEV heatmaps. The positive mask accumulates to exclude the easy positive candidates of prior stages from BEV heatmaps. The left object candidates are chosen and collected according to the response of BEV heatmap in a multi-stage process. (b) A deformable transformer decoder is adapted to effectively handle diverse object queries. The query embedding is enhanced with a box pooling module, which leverages the intermediate object supervision to identify local regions. It refines object queries in a local-scope manner, rather than at a point level. Residual connections and normalization layers have been excluded from the figure for clarity.}
	\label{fig:pipeline}
\end{figure*}

Based on HIP, we introduce a 3D object detector, FocalFormer3D, as shown in Fig.~\ref{fig:pipeline}. Especially, multi-stage heatmap predictions~\cite{centerpoint, centernet} are employed to excavate difficult instances. We maintain a class-aware \textit{Accumulated Positive Mask}, indicating positive regions from prior stages. Through this masking design, the model omits the training of easy positive candidates and thereby focuses on the hard instances (\textit{False Negatives}). Finally, our decoder collects the positive predictions from all stages to produce the object candidates. FocalFormer3D consistently demonstrates considerable gains over baselines in terms of average recall.

In addition, we also introduce a box-level refinement step to eliminate redundant object candidates. The approach employs a deformable transformer decoder~\cite{deformabledetr} and represents the candidates as box-level queries using RoIAlign. This allows for box-level query interaction and iterative box refinements, binding the object queries with sufficient box context through RoIAlign~\cite{maskrcnn, fasterrcnn} on the bird's eye view to perform relative bounding box refinements. Finally, a rescoring strategy is adopted to select positive objects from object candidates. Our ablation study in Table~\ref{tab: effects of restoring module} demonstrates the effectiveness of the local refinement approach in processing adequate object candidates.





Our contributions can be summarized as follows:
\begin{itemize}
    \item We propose Hard Instance Probing (HIP), a learnable scheme to automatically identify \textit{False Negatives} in a multi-stage manner. 
    \item We present FocalFormer3D for 3D object detection that effectively harvests hard instances on the BEV and demonstrates effectiveness in terms of average recall.
    \item Without bells and whistles, our model achieves state-of-the-art detection performance on \textbf{both} LiDAR-based and multi-modal settings. Notably, our model ranks \textbf{1st} places on \textbf{both} nuScenes 3D LiDAR detection and tracking leaderboard at time of submission.
\end{itemize}

\section{Related Work}
Modern 3D object detectors, either LiDAR-based~\cite{VoxelNet, pointpillar, second, centerpoint, PIXOR, pvrcnn, fastpointrcnn, voxelrcnn, pvrcnn++, centerformer, focalsconv, largekernel3d, chen2023voxenext}, or Camera-based~\cite{oftnet, chen2020dsgn, liftsplatshoot, li2022bevformer, zhang2022beverse, huang2021bevdet, liu2022petr, chen2022dsgn++}, or Multi-Modal~\cite{transfusion, bevfusion, bevfusionmit, li2022uvtr, li2022vff, 4dnet, li2022deepfusion, chen2022autoalignv2, mvp, futr3d, MV3D, MMF, contfuse} 3D object detectors generally rely on BEV view representation~\cite{li2022delving}. These methods adopt dense feature maps or dense anchors, for conducting object prediction in a bird's eye view (BEV) space. Among these methods, VoxelNet~\cite{second} as the pioneer works discretize point clouds into voxel representation and applies dense convolution to generate BEV heatmaps. SECOND~\cite{second} accelerates VoxelNet with 3D sparse convolution~\cite{sparseconv} to extract 3D features. Some Pillar-based detectors~\cite{pointpillar, PIXOR, shi2022pillarnet, ge2020afdet} collapse the height dimension and utilize 2D CNNs for efficient 3D detection. 

Different from dense detectors, point-based 3D detectors \cite{pointrcnn, 3dssd, std, votenet} directly process point clouds via PointNet~\cite{pointnet, pointnet++} and perform grouping or predictions on the sparse representations. Concerning involvement of neighborhood query on point clouds, it becomes time-consuming and unaffordable for large-scale point clouds. Concerning computation and spatial cost, another line of 3D detectors directly predicts objects on sparse point clouds to avoid dense feature construction. SST \cite{sst} applies sparse regional attention and avoids downsampling for small-object detection. FSD \cite{fan2022fully} instead further recognize instances directly on sparse representations obtained by SST \cite{sst} and SparseConv for long-range detection. 

Recent multi-modal detectors~\cite{bevfusion, bevfusionmit, li2022uvtr, li2022deepfusion, futr3d, pointaugmenting} follow the similar paradigm of BEV detectors and incorporate the multi-view image features by physical projection or learnable alignments between LiDAR and cameras. TransFusion~\cite{transfusion} applies cross attention to obtain image features for each object query. Despite various kinds of modal-specific voxel feature encoders, these detectors finally produce dense BEV features for classification and regression at the heatmap level.

\section{Methodology}

We introduce Hard Instance Probing (HIP) for automated identifying hard instances (\textit{False Negative}s) in Section~\ref{sec:hip}. We then present the implementations for the two main components of FocalFormer3D.
Section~\ref{sec:encoder} describes our multi-stage heatmap encoder that harvests the \textit{False Negative}s for producing high-recall initial object candidates following HIP. Section~\ref{sec:decoder} introduces a box-level deformable decoder network that further distinguishes objects from these candidates. 

\begin{figure}[t]
	\begin{center}
		\includegraphics[width=.9\linewidth]{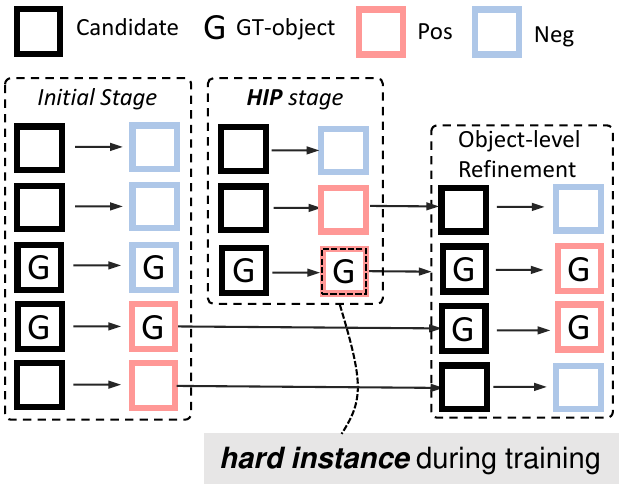}
	\end{center}
	\caption{\textbf{Hard Instance Probing.} We use the symbol ``G'' to indicate the object candidates that are labeled as ground-truth objects during the target assignment process in training. To ensure clarity, we omit numerous negative predictions for detection, given that background takes up most of the images.}
	\label{fig: toy example}
\end{figure}

\subsection{Hard Instance Probing (HIP)}
\label{sec:hip}
Real-world applications, such as autonomous driving, require a high level of scene understanding to ensure safe and secure operation. In particular, false negatives in object detection can present severe risks, emphasizing the need for high recall rates. However, accurately identifying objects in complex scenes or when occlusion occurs is challenging in 3D object detection, resulting in many false negative predictions. Unfortunately, few studies have explicitly focused on addressing false negatives in the design of detection heads. Motivated by the cascade-style detectors, we formulate a training pipeline to emulate the process of identifying false negative predictions at inference time.

\vspace{2mm}
\noindent\textbf{Formulation of Hard Instance Probing.}
Our strategy to identify hard instances operates stage by stage, as illustrated by a toy example in Fig.~\ref{fig: toy example}. Initially, we annotate the ground-truth objects as $$\mathcal{O} = \left\{o_i, i = 1, 2, ...\right\},$$ which is the main targets for initial stages. The neural network makes \textit{Positive} or \textit{Negative} predictions given a set of initial object candidates $\mathcal{A} = \left\{a_i, i = 1, 2, ... \right\}, $
which is not limited to anchors~\cite{fastrcnn}, point-based anchors~\cite{centerpoint}, and object queries~\cite{detr}. Suppose the detected objects (\textit{Positive} predictions) at $k$-th stage  are 
$$ \mathcal{P}_k = \left\{p_i, i = 1, 2, ...\right\}.$$ 
We are then allowed to classify the ground-truth objects according to their assigned candidates:
 $$\mathcal{O}_k^{TP} = \left\{o_j \big | \exists p_i\in \mathcal{P}_k, \sigma(p_i, o_j) > \eta \right\}.$$
where an object matching metric $\sigma(\cdot, \cdot)$ (e.g. Intersection over Union~\cite{kitti, waymo} and center distance~\cite{nuscenes}) and a predefined threshold $\eta$. Thus, the left unmatched targets can be regarded as hard instances:
$$ \mathcal{O}_k^{FN} = O - \bigcup_{i=1}^{k} O_k^{TP}. $$ 
The training of ($k+1$)-th stages is to detect these targets $\mathcal{O}_k^{FN}$ from the object candidates while omitting all prior  \textit{Positive} object candidates.

Despite the cascade way mimicking the process of identifying false negative samples, we might collect a number of object candidates across all stages. Thus, a second-stage object-level refinement model is necessary to eliminate any potential false positives.

\vspace{2mm}
\noindent\textbf{Relation with hard example mining.}
The most relevant topic close to our approach is hard example mining~\cite{exampleselection, ohem}, which samples hard examples during training. Recent research~\cite{focalloss, ghm, pisa} has further explored soft-sampling, such as adjusting the loss distribution to mitigate foreground-background imbalance issues. In contrast, our method operates in stages. Specifically, we use \textit{False Negative} predictions from prior stages to guide the subsequent stage of the model toward learning from these challenging objects. 


\subsection{Multi-stage Heatmap Encoder}
\label{sec:encoder}
The upcoming subsections outline the key implementations of FocalFormer3D as depicted in Fig.~\ref{fig:pipeline}. We begin by detailing the implementation of hard instance probing for BEV detection. This involves using the BEV center heatmap to generate the initial object candidate in a cascade manner.

\begin{figure}[t]
	\begin{center}
		\includegraphics[width=1.\linewidth]{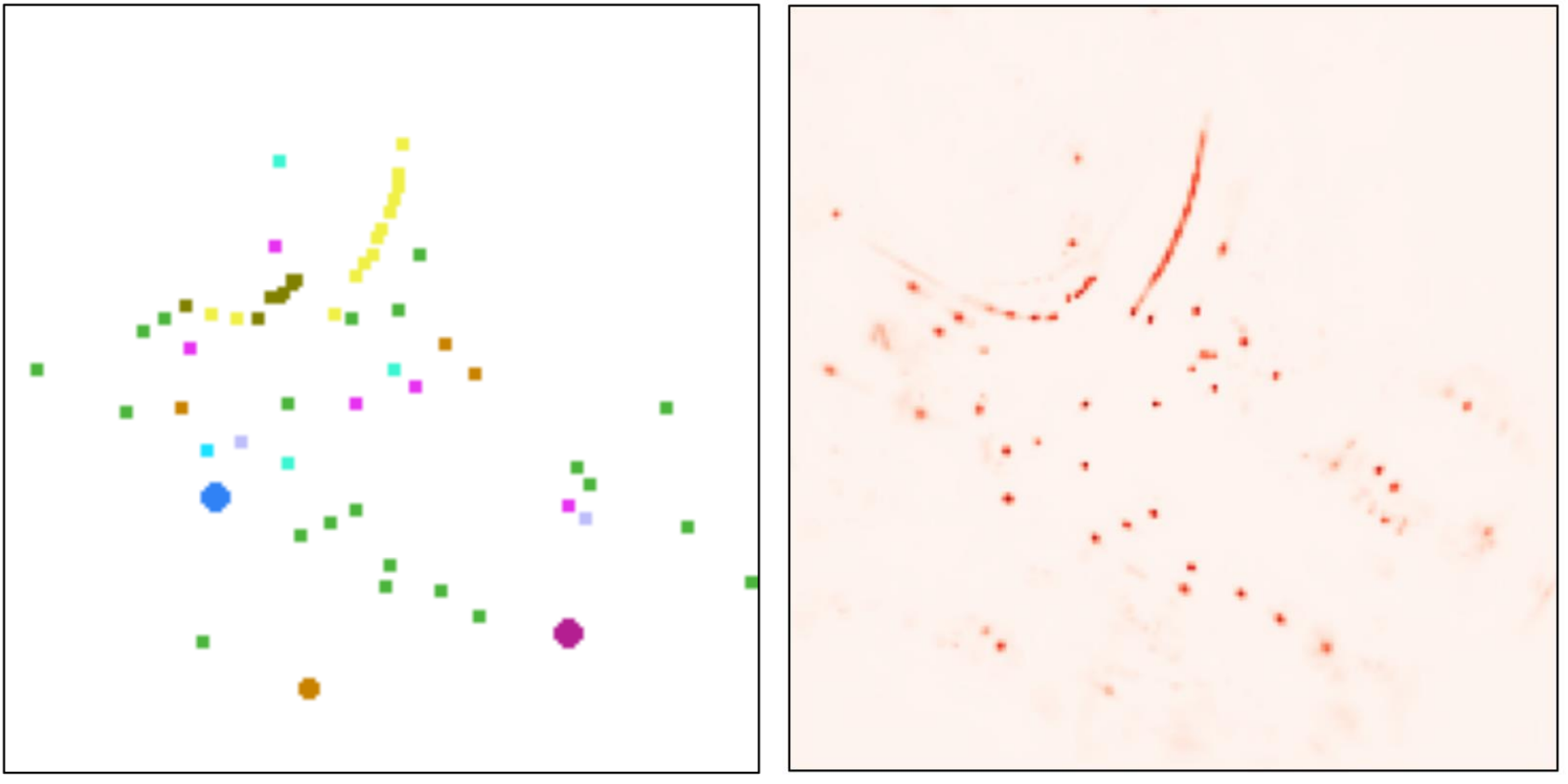}
	\end{center}
	\caption{\textbf{Example visualization for the positive mask. (left) and predicted BEV heatmap (right)}. The positive mask is class-aware and we show different categories with different colors for visualization. The masking area for objects of different categories can differ in the pooling-based masking method.}
	\label{fig:heatmap}
\end{figure}

\vspace{1mm}
\noindent\textbf{Preliminary of center heatmap in BEV perception.} 
In common practice~\cite{centerpoint, centernet, transfusion}, the objective of the BEV heatmap head is to produce heatmap peaks at the center locations of detected objects. The BEV heatmaps are represented by a tensor $S\in \mathbb{R}^{X\times Y\times C}$, where $X\times Y$ indicates the size of BEV feature map and $C$ is the number of object categories. The target is achieved by producing 2D Gaussians near the BEV object points, which are obtained by projecting 3D box centers onto the map view. In top views such as Fig.~\ref{fig:heatmap}, objects are more sparsely distributed than in a 2D image. Moreover, it is assumed that objects do not have intra-class overlaps on the bird's eye view.


Based on the non-overlapping assumption, excluding prior easy positive candidates from BEV heatmap predictions can be achieved easily. In the following, we illustrate the implementation details of HIP, which utilizes an accumulated positive mask.

\vspace{1mm}
\noindent\textbf{Positive mask accumulation.}
To keep track of all easy positive object candidates of prior stages, we generate a positive mask (PM) on the BEV space for each stage and accumulated them to an accumulated positive mask (APM):
$$\hat{M}_{k} \in \left\{0, 1\right\}^{X\times Y\times C},$$ which is initialized as all zeros. 

The generation of multi-stage BEV features is accomplished in a cascade manner using a lightweight inversed residual block~\cite{sandler2018mobilenetv2} between stages. Multi-stage BEV heatmaps are generated by adding an extra convolution layer. At each stage, we generate the positive mask according to the positive predictions. To emulate the process of identifying \textit{False Negative}s, we use a test-time selection strategy that ranks the scores according to BEV heatmap response~\cite{transfusion, centerpoint}. Specifically, at the $k$-th stage, Top-K selection is performed on the BEV heatmap across all BEV positions and categories, producing a set of object predictions $\mathcal{P}_k$. Then the positive mask $M_k\in \left\{0, 1\right\}^{X\times Y\times C}$ records the all the positions of positive predictions by setting $M_{(x,y,c)}=1$ for each predicted object $p_i\in \mathcal{P}_k$, where $(x, y)$ represents $p_i$'s location and $c$ is $p_i$'s class. The left points are set to $0$ by default.


According to the non-overlapping assumption, the ideal way to indicate the existence of a positive object candidate (represented as a point in the center heatmap) on the mask is by masking the box if there is a matched ground truth box. However, since the ground-truth boxes are not available at inference time, we propose the following masking methods during training:
\begin{itemize}
    \item \textbf{Point Masking}. This method involves no change, where only the center point of the positive candidates is filled.
    \item \textbf{Pooling-based Masking}. In this method, smaller objects fill in the center points while larger objects fill in with a kernel size of $3\times 3$.
    \item \textbf{Box Masking}. This method requires an additional box prediction branch and involves filling the internal region of the predicted BEV box.
\end{itemize}

The accumulated positive mask (APM) for the $k$-th stage is obtained by simply accumulating prior Positive Masks as follows:
$$ \hat{M}_{k} = \max_{1\leq i\leq k} M_i. $$
By masking the BEV heatmap $S_{k}$ with 
$$\hat{S}_k = S_k \cdot (1 - \hat{M}_k),$$
we omit prior easy positive regions in the current stage, thus enabling the model to focus on the false negative samples of the prior stage (hard instances). To train the multi-stage heatmap encoder, we adopt Gaussian Focal Loss~\cite{transfusion} as the training loss function. We sum up the BEV heatmap losses across stages to obtain the final heatmap loss.


During both training and inference, we collect the positive candidates from all stages as the object candidates for the second-stage rescoring as the potential false positive predictions. 

\vspace{1mm}
\noindent\textbf{Discussion on implementation validity for HIP.} 
Although the HIP strategy is simple, the masking way has two critical criteria that need to be met to ensure valid implementation of HIP: 
\begin{itemize} 
    \item Exclusion of prior positive object candidates at the current stage. 
    \item Avoidance of removal of potential real objects (false negatives). 
\end{itemize}
Point masking satisfies both requirements based on the following facts. As the Top-K selection is based on ranking predicted BEV heatmap scores, the hottest response points are automatically excluded when a point is masked. Besides, the design of a class-aware positive mask ensures that non-overlapping assumptions at the intra-class level on the BEV are met. 

However, the point masking strategy is less efficient as only one BEV object candidate is excluded for each positive prediction compared with the ideal masking with ground-truth box guidance. Therefore, there is a trade-off between the masking area and the validity of the exclusion operation. We compare all three strategies in Table~\ref{tab: whether apply mask} and pooling-based masking performs better than others.




\subsection{Box-level Deformable Decoder}
\label{sec:decoder}
The object candidates obtained from the multi-stage heatmap encoder can be treated as positional object queries~\cite{efficientdetr, transfusion}. The recall of initial candidates improves with an increase in the number of collected candidates. However, redundant candidates introduce false positives, thereby necessitating a high level of performance for the following object-level refinement blocks. 

To enhance the efficiency of object query processing, we employ deformable attention~\cite{deformabledetr} instead of computationally intensive modules such as cross attention~\cite{detr} or box attention~\cite{nguyen2022boxer}. Unlike previous methods that used center point features as the query embedding~\cite{transfusion, efficientdetr}, we model the object candidates as box-level queries. Specifically, Specifically, we introduce object supervision between deformable decoder layers, facilitating relative box prediction.

\vspace{1mm}
\noindent\textbf{Box-pooling module.} To better model the relations between objects and local regions in the regular grid manner, we extract the box context information from the BEV features using simple RoIAlign~\cite{maskrcnn} in the Box-pooling module as Fig.~\ref{fig:pipeline}. In specific, given the intermediate predicted box, each object query extracts $7\times 7$ feature grid points ~\cite{maskrcnn} from the BEV map followed by two MLP layers. The positional encoding is also applied both for queries and all BEV points for extracting positional information. This allows us to update both the content and positional information into the query embedding. 
This lightweight module enhances the query feature for the deformable decoder (See Table~\ref{tab: effects of restoring module}).

\vspace{1mm}
\noindent\textbf{Decoder implementation.}
Following Deformable DETR~\cite{deformabledetr}, our model employs 8 heads in all attention modules, including multi-head attention and multi-head deformable attention. The deformable attention utilizes 4 sampling points across 3 scales. To generate three scales of BEV features, we apply $2\times$ and $4\times$ downsampling operations to the original BEV features. The box-pooling module extracts $7\times 7$ feature grid points within each \textit{rotated} BEV box followed by 2 FC layers and adds the object feature to query embedding. We expand the predicted box to $1.2\times$ size of its original size. 



\begin{table*}[bpt]
	\begin{center}
		\begin{adjustbox}{width=1.\textwidth}
		\begin{tabular}{lccccccccccccc}
			\toprule
			Methods & Modality & mAP & NDS & Car & Truck & C.V. & Bus & Trailer & Barrier & Motor. & Bike & Ped. & T.C. \\ \midrule \midrule
                \multicolumn{5}{l}{\textit{LiDAR-based 3D Detection}} \\
                \midrule
			PointPillars \cite{pointpillar} & L & 30.5 & 45.3 & 68.4 & 23.0 & 4.1 & 28.2 & 23.4 & 38.9 & 27.4 & 1.1 & 59.7 & 30.8 \\
			CBGS \cite{cbgs} & L & 52.8 & 63.3 & 81.1 & 48.5 & 10.5 & 54.9 & 42.9 & 65.7 & 51.5 & 22.3 & 80.1 & 70.9 \\
			LargeKernel3D \cite{largekernel3d} & L & 65.3 & 70.5 & 85.9 & 55.3 & 26.8 & 66.2 & 60.2 & 74.3 & 72.5 & 46.6 & 85.6 & 80.0 \\
			TransFusion-L \cite{transfusion} & L & 65.5 & 70.2 & 86.2 & 56.7 & 28.2 & 66.3 & 58.8 & 78.2 & 68.3 & 44.2 & 86.1 & 82.0 \\
                PillarNet-34~\cite{shi2022pillarnet} & L & 66.0 & 71.4 & 87.6 &\textbf{57.5} & 27.9 & 63.6 & 63.1 & 77.2 & 70.1 & 42.3 & 87.3 & 83.3 \\
            LiDARMultiNet~\ \cite{ye2022lidarmultinet} & L & 67.0 & 71.6 & 86.9 & 57.4 & 31.5 & 64.7 & 61.0 & 73.5 & 75.3 & 47.6 & 87.2 & \textbf{85.1} \\ 
			\textbf{FocalFormer3D} & L & \textbf{68.7} & \textbf{72.6} & \textbf{87.2} & 57.1 & \textbf{34.4} & \textbf{69.6} & \textbf{64.9} & 
            \textbf{77.8} & \textbf{76.2} & \textbf{49.6} & \textbf{88.2} & 82.3
                 \\ \midrule
			CenterPoint \cite{centerpoint} $^\dag$ & L & 60.3 & 67.3 & 85.2 & 53.5 & 20.0 & 63.6 & 56.0 & 71.1 & 59.5 & 30.7 & 84.6 & 78.4 \\
                MGTANet$^\dag$~\cite{mgtanet} & L & 67.5 & 72.7 & 88.5 & 59.8 & 30.6 & 67.2 & 61.5 & 66.3 & 75.8 & 52.5 & 87.3 & \textbf{85.5} \\                 LargeKernel3D$^\ddag$~\cite{largekernel3d} & L & 68.8 & 72.8 & 87.3 & 59.1 & 30.2 & 68.5 & 65.6 & 75.0 & \textbf{77.8} & \textbf{53.5} & 88.3 & 82.4 \\
                \textbf{FocalFormer3D} $^\dag$ & L & \textbf{70.5} & \textbf{73.9} & \textbf{87.8} & \textbf{59.4} & \textbf{37.8} & \textbf{73.0} & \textbf{65.7} & \textbf{77.8} & 77.4 & 52.4 & \textbf{90.0} & 83.4 \\ 
			\midrule \midrule
                \multicolumn{5}{l}{\textit{Multi-Modal 3D Detection}} \\
                \midrule
			PointPainting \cite{pointpainting} & L+C & 46.4 & 58.1 & 77.9 & 35.8 & 15.8 & 36.2 & 37.3 & 60.2 & 41.5 & 24.1 & 73.3 & 62.4 \\
			3D-CVF \cite{3dcvf} & L+C & 52.7 & 62.3 & 83.0 & 45.0 & 15.9 & 48.8 & 49.6 & 65.9 & 51.2 & 30.4 & 74.2 & 62.9 \\
			MVP \cite{mvp} & L+C & 66.4 & 70.5 & 86.8 & 58.5 & 26.1 & 67.4 & 57.3 & 74.8 & 70.0 & 49.3 & 89.1 & 85.0 \\
			FusionPainting \cite{Fusionpainting} & L+C & 68.1 & 71.6 & 87.1 & 60.8 & 30.0 & 68.5 & 61.7 & 71.8 & 74.7 & 53.5 & 88.3 & 85.0 \\
			TransFusion \cite{transfusion} & L+C & 68.9 & 71.7 & 87.1 & 60.0 & 33.1 & 68.3 & 60.8 & 78.1 & 73.6 & 52.9 & 88.4 & 86.7 \\
			BEVFusion \cite{bevfusion} & L+C & 69.2 & 71.8 & 88.1 & 60.9 & 34.4 & 69.3 & 62.1 & 78.2 & 72.2 & 52.2 & 89.2 & 85.2 \\
			BEVFusion-MIT \cite{bevfusionmit} & L+C & 70.2 & 72.9 & \textbf{88.6} & 60.1 & \textbf{39.3} & 69.8 & 63.8 & 80.0 & 74.1 & 51.0 & 89.2 & 86.5 \\
			DeepInteraction \cite{deepinteraction} & L+C & 70.8 & 73.4 & 87.9 & 60.2 & 37.5 & 70.8 & 63.8 & \textbf{80.4} & 75.4 & 54.5 & \textbf{91.7} & \textbf{87.2} \\
			\textbf{FocalFormer3D} & L+C & \textbf{71.6} & \textbf{73.9} & 88.5 & \textbf{61.4} & 35.9 & \textbf{71.7} & \textbf{66.4} & 79.3 & \textbf{80.3} & \textbf{57.1} & 89.7 & 85.3 \\
			\midrule
			PointAugmenting \cite{pointaugmenting} $^\dag$ & L+C & 66.8 & 71.0 & 87.5 & 57.3 & 28.0 & 65.2 & 60.7 & 72.6 & 74.3 & 50.9 & 87.9 & 83.6 \\
			Focals Conv-F \cite{focalsconv} $^\ddag$ & L+C & 70.1 & 73.6 & 87.5 & 60.0 & 32.6 & 69.9 & 64.0 & 71.8 & 81.1 & 59.2 & 89.0 & 85.5 \\
			LargeKernel3D-F \cite{largekernel3d} $^\ddag$ & L+C & 71.1 & 74.2 & 88.1 & 60.3 & 34.3 & 69.1 & 66.5 & 75.5 & 82.0 & \textbf{60.3} & 89.6 & 85.7 \\
			\textbf{FocalFormer3D-F} $^\dag$  & L+C & \textbf{72.9} & \textbf{75.0} & \textbf{88.8} & \textbf{63.5} & \textbf{39.0} & \textbf{73.7} & \textbf{66.9} & \textbf{79.2} & \textbf{81.0} & 58.1 & \textbf{91.1} & \textbf{87.1} \\
			\bottomrule
		\end{tabular}
		\end{adjustbox}
	\end{center}
	\caption{\textbf{Performance comparison on the nuScenes 3D detection \textit{test} set.} $\dag$ represents using flipping test-time augmentation. $\ddag$ means using both flipping and rotation test-time augmentation. C.V, Motor., Ped. and T.C. are short for construction vehicle, motorcycle, pedestrian, and traffic cones, respectively.
 }
	\label{tab: nuScenes test results}
\end{table*}

\subsection{Model Training}
\label{sec:training}
The model is trained in two stages. In the first stage, we train the LiDAR backbone using a deformable transformer decoder head, which we refer to as DeformFormer3D (Table~\ref{tab: stages comparison} (a)).
After initializing the weights from DeformFormer3D, we train the FocalFormer3D detector, which consists of a multi-stage heatmap encoder and a box-level deformable decoder. However, during the training of the deformable decoder with bipartite graph matching, we encounter slow convergence issues in the early stages~\cite{dndetr}. To address this, we generate noisy queries from ground-truth objects~\cite{cmt, dndetr, dino}, enabling effective training of the model from scratch. Additionally, we improve the training process by excluding matching pairs with a center distance between the prediction and its GT object exceeding 7 meters.

\section{Experiments}

\subsection{Experimental Setup}

\vspace{1mm}
\noindent\textbf{Dataset and metric.} We evaluate our approach on \textit{nuScenes} and \textit{Waymo} 3D detection dataset.

\textit{nuScenes Dataset}~\cite{nuscenes} is a large-scale outdoor dataset. nuScenes contains $1,000$ scenes of multi-modal data, including $32$-beams LiDAR with $20$FPS and $6$-view camera images. We mainly evaluate our method on both \textit{LiDAR-only} and \textit{LiDAR-Camera fusion} settings. The evaluation metrics follow nuScenes official metrics including mean average precision (mAP) and nuScenes detection score (NDS) defined by averaging the matching thresholds of center distance $\mathbb{D} = \{0.5, 1., 2., 4.\}$ (m). For evaluating the quality of object queries, we also introduce the Average Recall (AR) defined by center distance as well. The ablation studies in our research primarily utilize the nuScenes dataset, unless explicitly stated otherwise.

\textit{Waymo Open Dataset}~\cite{waymo} has a wider detection range of $150m\times 150m$ compared to the nuScenes dataset. Waymo dataset comprises of 798 scenes for training and 202 scenes for validation. The official evaluation metrics used are mean Average Precision (mAP) and mean Average Precision with Heading (mAPH), where the mAP is weighted by the heading accuracy. The mAP and mAPH scores are computed with a 3D Intersection over Union (IoU) threshold of 0.7 for \textit{Vehicle} and 0.5 for \textit{Pedestrian} and \textit{Cyclist}. The evaluation has two difficulty levels: Level 1, for boxes with more than five LiDAR points, and Level 2, for boxes with at least one LiDAR point. Of the two difficulty levels, Level 2 is prioritized as the primary evaluation metric for all experiments. 

\vspace{1mm}
\noindent\textbf{Implementation details.}
Our implementation is mainly based on the open-sourced codebase MMDetection3D \cite{mmdet3d2020}. For the LiDAR backbone, we use CenterPoint-Voxel as the point cloud feature extractor. For the multi-stage heatmap encoder, we apply 3 stages, generating a total of 600 queries by default. Data augmentation includes random double flipping along both $X$ and $Y$ axes, random global rotation between [$-\pi/4$, $\pi/4$], the random scale of [$0.9$, $1.1$], and random translation with a standard deviation of $0.5$ in all axes. All models are trained with a batch size of 16 on eight V100 GPUs. More implementation details are referred to in supplementary files.


\subsection{Main Results}
\noindent\textbf{nuScenes LiDAR-based 3D object detection.}
We evaluate the performance of FocalFormer3D on the nuScenes \textit{test} set. As shown in Table \ref{tab: nuScenes test results}, the results demonstrate its superiority over state-of-the-art methods on various evaluation metrics and settings. Our single-model FocalFormer3D achieved $68.7$ mAP and $72.6$ NDS, which surpasses the prior TransFusion-L method by $+3.2$ points on mAP and $+2.4$ points on NDS. Notably, even compared with the previous best method that was trained with segmentation-level labels, our method without extra supervision still outperformed LiDARMultiNet by $+1.7$ mAP and $+1.0$ NDS. 

\vspace{2mm}
\noindent\textbf{nuScenes multi-modal 3D object detection.} We extend our approach to a simple multi-modal variant and demonstrate its generality. Following TransFusion~\cite{transfusion}, we use a pre-trained \textit{ResNet-50} model on COCO~\cite{coco} and nuImage~\cite{nuscenes} dataset as the image model and freeze its weights during training. To reduce computation costs, the input images are downscaled to 1/2 of their original size. Unlike heavy lift-splat-shot~\cite{liftsplatshoot} camera encoders used in BEVFusion~\cite{bevfusion, bevfusionmit}, the multi-view camera images are projected onto a pre-defined voxel space and fused with LiDAR BEV feature. Additional details are available in the supplementary files. Without test-time augmentation, our simple multi-modal variant model outperforms all other state-of-the-art with less inference time (Table~2). With TTA, FocalFormer3D achieves 72.9 mAP and 75.0 NDS, ranking first among all single-model solutions on the nuScenes benchmark. Interestingly, our model achieves high results for some rare classes such as (\textsl{Trailer}, \textsl{Motorcycle}, \textsl{Bicycle}) compared to other methods.

\vspace{2mm}
\noindent\textbf{nuScenes 3D object tracking.} 
To further demonstrate the versatility, we also extend FocalFormer3D to 3D multi-object tracking (MOT) by using the tracking-by-detection algorithm SimpleTrack. Interested readers can refer to the original paper~\cite{pang2021simpletrack} for more comprehensive details. As depicted in Table~\ref{tab: nuscenes tracking}, FocalFormer3D gets 2.9 points better than prior state-of-the-art TransFusion-L~\cite{transfusion} in LiDAR settings and FocalFormer3D-F achieves 2.1 points over TransFusion in terms of AMOTA. Moreover, our single model FocalFormer3D-F with double-flip testing results performs even better than the BEVFusion~\cite{bevfusionmit} with model ensembling.

\begin{table}[t]
	\begin{center}
        \resizebox{\linewidth}{!}{
		\begin{tabular}{lcccc}
			\toprule
			 Methods &  AMOTA   & AMOTP &  MOTA & IDS \\ \midrule \midrule
                \multicolumn{5}{l}{\textit{LiDAR-based 3D Tracking}} \\
                \midrule
                AB3DMOT~\cite{ab3dmot} & 15.1 & 150.1 & 15.4 & 9027 \\
                CenterPoint~\cite{centerpoint} & 63.8 & 55.5 & 53.7 & 760 \\
                CBMOT~\cite{cbmot} & 64.9 & 59.2 & 54.5 & 557 \\
                OGR3MOT~\cite{ogr3mot} & 65.6 & 62.0 & 55.4 & \textbf{288} \\             SimpleTrack~\cite{pang2021simpletrack} & 66.8 & 55.0 & 56.6 & 575 \\
                UVTR-L~\cite{li2022uvtr} & 67.0 & 55.0 & 56.6 & 774 \\
                TransFusion-L~\cite{transfusion} & 68.6 & 52.9 & 57.1 & 893 \\
                \midrule
                \textbf{FocalFormer3D}  & 71.5 & 54.9 & \textbf{60.1} & 888 \\
                \textbf{FocalFormer3D}$^\dag$  & \textbf{72.1} & \textbf{47.0} & 60.0 & 701 \\
                \midrule \midrule
                \multicolumn{5}{l}{\textit{Multi-Modal 3D Tracking}} \\
                \midrule
                UVTR-MultiModal~\cite{li2022uvtr} & 70.1 & 68.6 & 61.8 & 941 \\
                TransFusion~\cite{transfusion} & 71.8 & 55.1 & \textbf{60.7} & 944 \\
                BEVFusion-MIT~\cite{bevfusionmit}$^\ddag$ & 74.1 & 40.3 & 60.3 & 506 \\ 
                \midrule
                \textbf{FocalFormer3D-F} & 73.9 & 51.4 & 61.8 & \textbf{824} \\
                \textbf{FocalFormer3D-F}$^\dag$ & \textbf{74.6} & \textbf{47.3} & 63.0 & 849 \\
			\bottomrule
		\end{tabular}}
	\end{center}
	\caption{\textbf{Performance comparison on nuScenes 3D tracking test set}. $^{\dagger}$ is based on the double-flip testing results in Table~\ref{tab: nuScenes test results}. $^{\ddag}$ is based on model ensembling. } \label{tab: nuscenes tracking}
 \vspace{-1mm}
\end{table}

\vspace{2mm}
\noindent\textbf{Waymo LiDAR 3D object detection.}
The results of our single-frame LiDAR 3D detection method on the Waymo dataset are presented in Table~\ref{tab:waymo results}, alongside the comparison with other approaches. Employing with the same VoxelNet backbone as nuScenes, our method achieves competitive performance without any fine-tuning of the model hyperparameters specifically for the Waymo dataset. Particularly, when compared to TransFusion-L with the same backbone, our method exhibits a +1.1 mAPH improvement.

\begin{table}[t]
	\begin{center}
         \resizebox{\linewidth}{!}{
		\begin{tabular}{lccccc}
		\toprule
		  Methods & mAP & mAPH & Vel. & Ped. & Cyc.  \\ \midrule\midrule
                \multicolumn{5}{l}{\textit{LiDAR-based 3D Detection}} \\ \midrule
            RSN$^{\star}$~\cite{RSN} & -- & -- & 65.5 & 63.7 & --  \\
            AFDetV2$^{\star}$~\cite{hu2022afdetv2} & 71.0 & 68.8 & 69.2 & 67.0 & 70.1 \\
            SST$^{\star}$~\cite{sst} & 67.8 & 64.6 & 65.1 & 61.7 & 66.9 \\
            PV-RCNN$^{\star}$~\cite{pvrcnn} & 66.8 & 63.3 & 68.4 & 65.8 & 68.5 \\
            PV-RCNN++$^{\star}$~\cite{pvrcnn++} & 71.7 & 69.5 & 70.2 & \textbf{68.0} & 70.2 \\
            PillarNet-34$^{\star}$~\cite{shi2022pillarnet} & 71.0 & 68.8 & \textbf{70.5} & 66.2 & 68.7 \\
            FSD-spconv$^{\star}$~\cite{fan2022fully} & \textbf{71.9} & \textbf{69.7} & 68.5 & \textbf{68.0} & 72.5 \\ \midrule
            CenterPoint~\cite{centerpoint} & 69.8 & 67.6 & 73.4 & 65.8 & 68.5 \\
            TransFusion-L$^\wedge$~\cite{transfusion} & 70.5 & 67.9 & 66.8 & 66.1 & 70.9 \\
            FocalFormer3D & 71.5 & 69.0 & 67.6 & 66.8 & \textbf{72.6} \\
            \bottomrule
		\end{tabular}
        }
	\end{center}
	\caption{\textbf{Performance comparison on the Waymo \textit{val} set.} All models inputs single-frame point clouds. The methods marked with $^*$ indicate the utilization of different point cloud backbones in VoxelNet. The method marked with $^\wedge$ indicates our reproduction. The evaluation metric used is the LEVEL 2 difficulty, and the results are reported on the full Waymo validation set.}\label{tab:waymo results}
\end{table}

\subsection{Recall Analysis}

\begin{figure*}[t]
	\begin{center}
		\includegraphics[width=1.\linewidth]{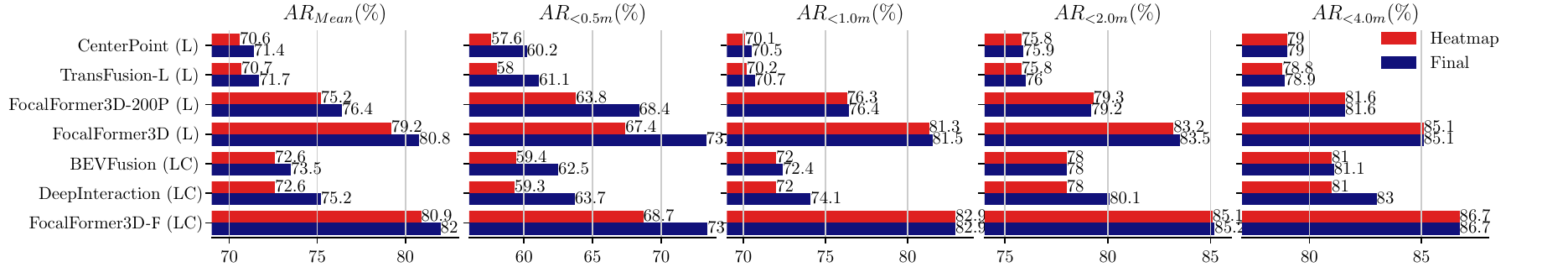}
	\end{center}
	\caption{Average recall comparisons between initial object predictions and final object prediction centers on the nuScenes \textit{val} set. The subfigures are shown over center distance thresholds (\%) following nuScenes detection metrics. }
	\label{fig:recall}
\end{figure*}

To diagnose the performance improvements, we compare several recent methods in terms of AR for both stages -- initial BEV heatmap predictions and final box predictions in Fig.~\ref{fig:recall}. The metric of AR is computed based on center distance following the nuScenes metrics and different distance thresholds (\eg, $0.5m$, $1.0m$, $2.0m$, $4.0m$), and the mean AR (mAR) are compared. 

\begin{figure}[t]
	\begin{center}
		\includegraphics[width=1.\linewidth]{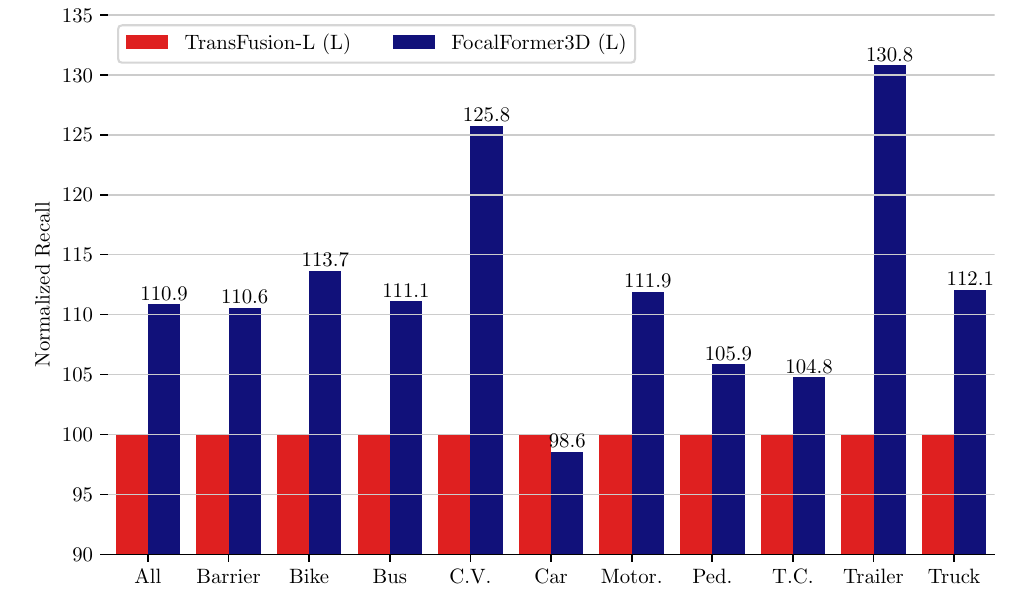}
	\end{center}
	\caption{\textbf{Class-wise recall comparison on nuScenes val set} between TransFusion-L and FocalFormer3D in terms of recall values across nuScenes center distance (CD) threshes (0.25/0.5/1.0m) on the nuScenes \textit{val} set. The red bars are normalized to 100\%.}
	\label{fig:classwise recall}
\end{figure}

\vspace{2mm}
\noindent\textbf{Recall comparison on initial object candidates.}
Figure \ref{fig:recall} compares the recall of state-of-the-art methods that share the same SparseUNet backbone. With total 200 queries, FocalFormer3D-200P reaches 75.2 mAR, achieving considerable and consistent improvements by $+4.5$ mAR compared with the prior state-of-the-art LiDAR approach TransFusion-L. Surprisingly, our LiDAR-based FocalFormer even achieves better results than the prior multi-modal approach DeepInteraction by 2.6 points in terms of mAR as well. As the query sizes get 600, FocalFormer3D achieves 79.2 mAR, surpassing the fusion approach DeepInteraction by 6.6 points. Further, by incorporating multi-view camera features, our multi-modal version FocalFormer-F gets improved to 80.9 mAR.

\vspace{2mm}
\noindent\textbf{Recall comparison on final object prediction.}
Concerning the final predictions of 3D detectors, most LiDAR and fusion approaches obtain fewer performance improvements as the distance thresholds increase as shown in Fig.~\ref{fig:recall}. This can be explained by higher distance thresholds indicating the performance for the extreme cases of missing detections. The introduction of camera features helps the model see the context in the perspective view, which leads to better performance such as DeepInteraction. However, their final prediction recall falls far behind FocalFormer-F with a large margin of 6.8 points. 

\vspace{2mm}
\noindent\textbf{Class-wise recall comparison.}
We compare the class-wise recall analysis for object candidates in Fig.~\ref{fig:classwise recall} at the category level. The findings highlight the effectiveness of FocalFormer3D in improving the relative recall of initial BEV queries by a relative +10.9\% improvement against TransFusion-L. Large objects such as \textsl{Construction Vehicles} and \textsl{Trailer} get the most improvements so that the predictions of their initial centers are challenging.

\subsection{Ablation Study}

\vspace{2mm}
\noindent\textbf{HIP query sizes and generation stages.}
Table~\ref{tab: stages comparison} ablates the impacts of the number of queries and stages in the multi-stage heatmap encoder. When using the same query size of rough 200, approaches (b), which uses additional one stage of HIP, demonstrates better performance than baseline (a) by a margin of $+0.7$ mAP. When provided with more queries (600), our approach (d) and (e) achieve over 1.1-point improvement in terms of mAP. 

\begin{table}[bpt]
	\begin{center}
		\begin{tabular}{ccccc}
			\toprule
		\# & \# Stages & \#Total Queries  & mAP & NDS \\ \midrule
            (a) & 1 & 200 & 65.3 & 70.5 \\
            (b) & 2 & 200 & 66.0 & 70.7 \\
            (c) & 1 & 600 & 65.4 & 70.5 \\
            (d) & 2 & 600 & 66.4 & 70.9 \\
            (e) & 3 & 600 & 66.5 & 71.1 \\
			\bottomrule
		\end{tabular}
	\end{center}
	\caption{\textbf{Effects of numbers of stages and total queries.} Here one stage stands for the baseline method without using hard instance probing. } \label{tab: stages comparison}
\end{table}

\begin{table}[bpt]
	\begin{center}
		\begin{tabular}{lcc}
			\toprule
			  Mask Type & mAP & NDS \\ \midrule
                None  & 65.3 & 70.4 \\ \midrule
                Point-based & 65.9 & 70.5 \\
                Box-based & 66.1 & 70.9 \\ 
                Pooling-based & 66.5 & 71.1 \\
			\bottomrule
		\end{tabular}
	\end{center}
	\caption{\textbf{Effects of various positive mask types.} All models adopt the same network except for the masking way.} \label{tab: whether apply mask}
\end{table}

\vspace{2mm}
\noindent\textbf{Positive mask type.}
Table~\ref{tab: whether apply mask} presents an ablation study on the effectiveness of Hard Instance Probing in terms of various mask types. Specifically, we compare the performance of our method with none masking, point-based masking, and pooling-based masking. The results demonstrate that even with single-point masking, HIP improves the performance of the baseline by a gain of $+0.6$ points in terms of mAP. Furthermore, the pooling-based masking shows the best gain with $+1.2$ mAP and $+0.7$ NDS, outperforming the box-based masking. This can be attributed to two facts. Point or pooling-based masking can already effectively exclude positive objects as the center heatmap~\cite{centerpoint} only highlights a Gaussian peak. Second, the wrong false positive predictions or predicted boxes might lead to false masking of the ground-truth boxes, resulting in missed detection.

\begin{table}[bpt]
	\begin{center}
    \resizebox{\linewidth}{!}{
		\begin{tabular}{ccccccc}
			\toprule
                \multirow{2}{*}{\#} & \multirow{2}{*}{M.S. Heat} & \multicolumn{2}{c}{Refinement Module} & \multirow{2}{*}{mAP} & \multirow{2}{*}{NDS} \\ \cmidrule(lr){3-4}
                & & BoxPool & C.A. &  & \\
                \midrule
    		(a) & \xmark & \xmark & \xmark & 63.1 & 69.1 \\ 
                (b) & \cmark & \xmark & \xmark & 63.3 & 69.3 \\ 
                (c) & \cmark & \cmark & \xmark & 65.1 & 69.9 \\ 
                (d) & \cmark & \xmark & \cmark & 65.9 & 70.9 \\ 
                (e) & \cmark & \cmark & \cmark & 66.5 & 71.1 \\ \midrule
                (f) & \cmark & \multicolumn{2}{c}{Rescoring Only} & 66.1 & 68.8 \\
			\bottomrule
		\end{tabular}
  }
	\end{center}
	\caption{\textbf{Step-by-step improvements made by modules.} ``M.S. Heat'' represents the application of the multi-stage heatmap encoder for hard instance probing. ``C.A.'' denotes using deformable cross attention for second-stage refinement. ``BoxPool'' represents the Box-pooling module. The term ``Rescoring Only'' refers to the model that directly generates box prediction from BEV feature and uses its decoder head to rescore the candidate predictions from heatmap without performing additional bounding box refinement. } \label{tab: effects of restoring module}
\end{table}

\vspace{2mm}
\noindent\textbf{Step-by-step module refinement.}
We conduct ablation studies on the step-by-step improvements by each module, presented in Table~\ref{tab: effects of restoring module}, to illustrate the component effectiveness within hard instance probing (HIP) pipeline. Initially, without second-stage refinement, we used simple center-based predictions~\cite{centerpoint} (a), which estimate boxes directly from BEV feature by another convolutional layer. 

Despite an improvement in the average recall by over $9$ points in Fig.~{\color{red} 5}, we found little improvement of (b) over (a) in performance after using the multi-stage heatmap encoder to generate the object candidates. By applying simple object-level rescoring (c), with RoI-based refinement (using two hidden MLP layers), the performance is boosted to $65.1$ mAP and $69.9$ NDS. Remarkably, our complete box-level deformable decoder (e) further improves the performance by a margin of $+1.4$ mAP and $+1.2$ NDS. 

To assess the effects of rescoring alone, we perform experiment (f), which excludes the effects of box regression by not using any box or position regression in the object-level refinement module. Despite this, experiment (f) still achieves high center accuracy ($66.1$ mAP) compared to (a). This finding highlights the limitations of the initial ranking of object candidates across stages based solely on BEV heatmap scores. Therefore, it validates the necessity for a second-stage object-level rescoring in the hard instance probing pipeline (Fig.~{\color{red} 3}). 


\vspace{2mm}
\noindent\textbf{Latency analysis for model components.}
We conduct a latency analysis for FocalFormer3D on the nuScenes dataset. The runtimes are measured on the same V100 GPU machine for comparison.To ensure a fair speed comparison with CenterPoint~\cite{centerpoint}, dynamic voxelization~\cite{dynamicvoxelization} is employed for speed testing of both TransFusion-L and FocalFormer3D. The computation time is mostly taken up by the sparse convolution-based backbone network (VoxelNet~\cite{VoxelNet, second}), which takes 78ms. Our multi-stage heatmap encoder takes 13ms to collect queries from the heatmaps across stages, while the box-level deformable decoder head takes 18ms. Note that, the generation of multi-stage heatmaps only takes 5ms, and additional operations such as Top-K selection takes 7ms, indicating potential optimization opportunities for future work.
\begin{table}[bpt]
	\begin{center}
		\begin{tabular}{lccccc}
		\toprule
		 Models/Components & Latency  \\ \midrule
            TransFusion-L & 93ms \\ \midrule
            FocalFormer3D & 109ms \\ 
            --   VoxelNet backbone & 78ms \\
            --   Multi-stage heatmap encoder & 13ms \\
            --   Box-level deformable decoder & 18ms \\
            \bottomrule
		\end{tabular}
	\end{center}
	\caption{\textbf{Latency analysis for model components.} Latency is measured on a V100 GPU for reference. }\label{tab:latency comparison}
\end{table}

\section{Conclusion}
In this work, we explicitly focus on the fatal problem in autonomous driving, {\em i.e.}, false negative detections. We present FocalFormer3D as solution. It progressively probes hard instances and improves prediction recall, via the hard instance probing~(HIP). Nontrivial improvements are introduced with limited overhead upon transformer-based 3D detectors. The HIP algorithm enables FocalFormer3D to effectively reduce false negatives in 3D object detection.
\paragraph{Limitation.}~\label{limitation and future work}
A key limitation is that FocalFormer3D's hard instance probing (HIP) relies on the assumption that object centers produce Gaussian-like peaks in the BEV heatmap, which may not hold for camera-based detectors where heatmaps tend to be fan-shaped. Additionally, few studies have explored hard instances in long-range detection, so more research is needed to evaluate HIP in this area. We leave more investigation of hard instance probing as future work.

{\small
\bibliographystyle{unsrt}
\bibliography{egbib}
}

\newpage
\appendix

\section*{Appendix for FocalFormer3D}

The supplementary materials for FocalFormer3D is organized as follows:
\begin{itemize}
    
    \item Sec.~\ref{sec:Additional ablation studies} shows additional ablation studies on decoder head and latency analysis for multi-modal models.
    \item Sec.~\ref{sec:implementation-details} gives more implementation details including network details, and extension to the multi-modal variant.
    \item Sec.~\ref{sec: prediction locality} discusses the \textit{prediction locality} for second-stage refinements.
    \item Sec.~\ref{sec: visualization} presents some visual results for multi-stage heatmaps and 3D detection results on bird's eye view.
\end{itemize}



\section{Additional Ablation Studies}
\label{sec:Additional ablation studies}

\vspace{2mm}
\noindent\textbf{Design of the decoder head.} We analyze the capability of the decoder head in processing massive queries in Table~\ref{tab: decoder head}. Concerning the type of cross attention, with an increasing number of queries up to 600, the computation time of cross attention module~\cite{detr} (c) grows faster than deformable one~\cite{deformabledetr} (e). As a result, more deformable transformer layers can be applied. In our experiments, the transformer decoder head with 6 layers obtains the best performance (66.5 mAP and 71.1 NDS) with a more affordable computation time than the cross attention modules. Furthermore, compared with point-level query embedding~\cite{transfusion} (g), our box-level query embedding (f) achieves +0.6 points improvements with $3.7ms$ computation overhead, demonstrating the effectiveness of box-level query.

\begin{table}[h]
	\begin{center}
    \resizebox{\linewidth}{!}{
		\begin{tabular}{ccccccc}
			\toprule
		\#	& C.A. & \#Q & \#Layer & mAP & NDS & Latency \\ \midrule
            (a) &  Full & 200 & 1 & 65.8 & 70.5 & 7.6ms \\
            (b) &  Full & 600 & 1 & 66.1 & 70.9 & 13.1ms \\
            (c) &  Full & 600 & 2 & 66.3 & 71.1 & 26.2ms \\
            (d) &  Deform & 200 & 6 & 65.9 & 70.8 & 14.8ms \\
            (e) &  Deform & 600 & 2 & 66.2 & 70.7 & 7.6ms \\ 
            (f) & Deform & 600 & 6 & 66.5 & 71.1 & 17.0ms \\ \midrule
            (g) & \multicolumn{3}{c}{w/o Box-pooling} & 65.9 & 70.9 & -- \\  
			\bottomrule
		\end{tabular}}
	\end{center}
	\caption{\textbf{Ablation studies for box-level deformable decoder head.} ``C.A.'' denotes the types of cross attention layers. ``\# Q'' represents the number of used queries. ``\# Layer'' stands for the number of decoder layers. Latency is measured for the transformer decoder head on a V100 GPU for reference.} \label{tab: decoder head}
\end{table}

\vspace{2mm}
\noindent\textbf{Latency analysis.}
We compare ours with other leading-performance methods in Table~\ref{tab:latency}. It shows that FocalFormer-F outperforms the dominating methods, BEVFusion~\cite{bevfusion} and DeepInteraction~\cite{deepinteraction} in terms of both performance and efficiency.

\begin{table}[h]
	\begin{center}
		\begin{tabular}{lccc}
		\toprule
		  Methods & mAP & NDS & Latency \\ \midrule
            BEVFusion~\cite{bevfusion} & 69.2 & 71.8 & 1610ms \\
            DeepInteraction~\cite{deepinteraction} &  70.8 & 73.4 & 480ms \\
            FocalFormer3D-F (Ours) & \textbf{71.6} & \textbf{73.9} & \textbf{363ms} \\ 
		\bottomrule
		\end{tabular}
	\end{center}
	\caption{\textbf{Efficiency comparison with other SOTA methods on nuScenes dataset.} Results are shown on nuScenes test set. All methods are tested on a single V100 GPU for reference.}\label{tab:latency}
\end{table}

\vspace{2mm}
\noindent\textbf{Results on nuScenes val set.}
We also report the method comparisons on the nuScenes \textit{val} set in Table~\ref{tab:nuscenes val result}. 

\begin{table}[h]
	\begin{center}
		\begin{tabular}{lccccc}
		\toprule
		  Methods & mAP & NDS  \\ \midrule
            CBGS~\cite{cbgs} & 51.4 & 62.6 \\
            CenterPoint~\cite{centerpoint} & 59.6 & 66.8 \\
            LiDARMultiNet~\cite{ye2022lidarmultinet} & 63.8 & 69.5 \\
            TransFusion-L$^\wedge$~\cite{transfusion} & 65.2 & 70.2 \\ 
            FocalFormer3D (Ours) & 66.5 & 71.1 \\
            \bottomrule
		\end{tabular}
	\end{center}
	\caption{\textbf{Performance comparison on the nuScenes \textit{val} set.} Results marked with $^\wedge$ indicate our reproduction. The results of other compared methods on the nuScenes \textit{val} set were obtained from their respective original papers.}\label{tab:nuscenes val result}
\end{table}

\section{Additional Implementation Details}
\label{sec:implementation-details}

\vspace{2mm}
\noindent\textbf{Model details for nuScenes dataset.}
On the nuScenes dataset, the voxel size is set as $0.075m\times 0.075m\times 0.2m$ and the detection range is set to [$-54.0m$, $54.0m$] along $X$ and $Y$ axes, and [$-5.0m$, $3.0m$] along $Z$ axis.We follow the common practice of accumulating the past 9 frames to the current frame for both training and validation. We train the LiDAR backbone with the deformable transformer decoder head for 20 epochs. Then, we freeze the pre-trained LiDAR backbones and train the detection head with multi-stage focal heatmaps for another 6 epochs. GT sample augmentation is adopted except for the last 5 epochs. We adopt pooling-based masking for generating Accumulated Positive Mask, where we simply select \textsl{Pedestrian} and \textsl{Traffic Cones} as the small objects. 

\vspace{2mm}
\noindent\textbf{Model details for Waymo dataset.}
On the Waymo dataset, we simply keep the VoxelNet backbone and FocalFormer3D detector head the same as those used for the nuScenes dataset. The voxel size used for the Waymo dataset is set to $0.1m \times 0.1m \times 0.15m$. For the multi-stage heatmap encoder, we use pooling-based masking, selecting \textsl{Vehicle} as the large object category, and \textsl{Pedestrain} and \textsl{Cyclist} as the small object categories. The training process involves two stages, with the model trained for 36 epochs and another 11 epochs trained for the FocalFormer3D detector. We adopt GT sample augmentation during training, except for the last 6 epochs. As the Waymo dataset provides denser point clouds than nuScenes, the models adopt single-frame point cloud input~\cite{centerpoint, transfusion}.

\vspace{2mm}
\noindent\textbf{Extension to multi-modal fusion model.}
We provide more details on the extension of FocalFormer3D to its multi-modal variant. Specifically, the image backbone network utilized is ResNet-50 following TransFusion~\cite{transfusion}. Rather than using more heavy camera projection techniques such as Lift-split-shot~\cite{liftsplatshoot} or BEVFormer~\cite{li2022bevformer}, we project multi-view camera features onto a predefined voxel grid in the 3D space~\cite{oftnet}. The BEV size of the voxel grid is set to $180\times 180$, in line with $8\times$ downsampled BEV features produced by VoxelNet~\cite{VoxelNet}. The height of the voxel grid is fixed at 10.

To obtain camera features for BEV LiDAR feature, we adopt a cross-attention module~\cite{deepinteraction} within each pillar. This module views each BEV pixel feature as the query and the projected camera grid features as both the key and value. The generated camera BEV features are then fused with LiDAR BEV features by an extra convolutional layer. This multi-modal fusion is conducted at each stage for the multi-stage heatmap encoder. We leave the exploration of stronger fusion techniques~\cite{bevfusion, deepinteraction, bevfusionmit} as future work.

\section{Prediction Locality of Second-Stage Refinement}
\label{sec: prediction locality}
\begin{figure}[h]
	\begin{center}
		\includegraphics[width=1.\linewidth]{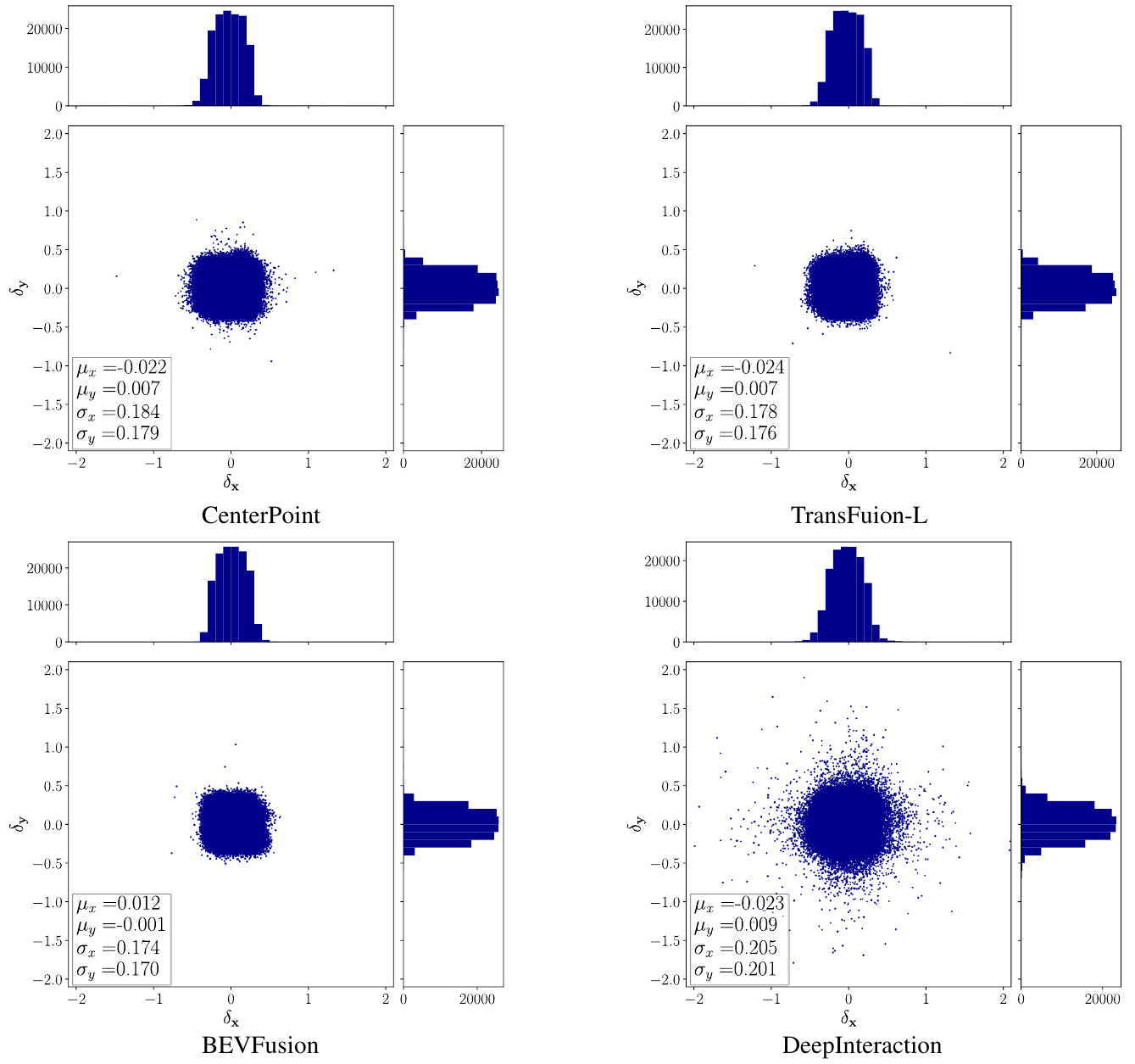}
	\end{center}
	\caption{Object center shifts $(\delta_x, \delta_y)$ distribution without normalization between initial heatmap response and final object predictions. The unit is a meter.}
	\label{fig:deltacenter}
\end{figure}

Recent 3D detectors have implemented global attention modules~\cite{transfusion} or fusion with multi-view camera~\cite{bevfusion, deepinteraction} to capture larger context information and improve the detection accuracy. However, we observe a limited regression range (named as \textit{prediction locality}) compared to the initial heatmap prediction. To analyze their second-stage ability to compensate for the missing detection (false negatives), we visualize the distribution of their predicted center shifts $\delta = (\delta_x, \delta_y)$ in Fig.~\ref{fig:deltacenter} for several recent leading 3D detectors, including the LiDAR detectors (CenterPoint~\cite{centerpoint}, TransFusion-L~\cite{transfusion}) and multi-modal detectors (BEVFusion~\cite{bevfusion}, DeepInteraction~\cite{deepinteraction}). Statistics of center shift ($\sigma_\delta < 0.283m$ illustrate almost all predictions are strongly correlated with their initial positions (generally less than $2$ meters away), especially for LiDAR-only detectors, such as CenterPoint and TransFusion-L. 

The disparity between small object sizes (usually $<5m\times 5m$) and extensive detection range (over $100m\times 100m$ meters) limits the efficacy of long-range second-stage refinement, despite the introduction of global operations and perspective camera information. Achieving a balance between long-range modeling and computation efficiency for BEV detection is crucial. FocalFormer3D, as the pioneer in identifying false negatives on the BEV heatmap followed by local-scope rescoring, may provide insights for future network design.


\section{Example Visualization}
\label{sec: visualization}

\begin{figure*}[t]
	\begin{center}
		\includegraphics[width=0.75\linewidth]{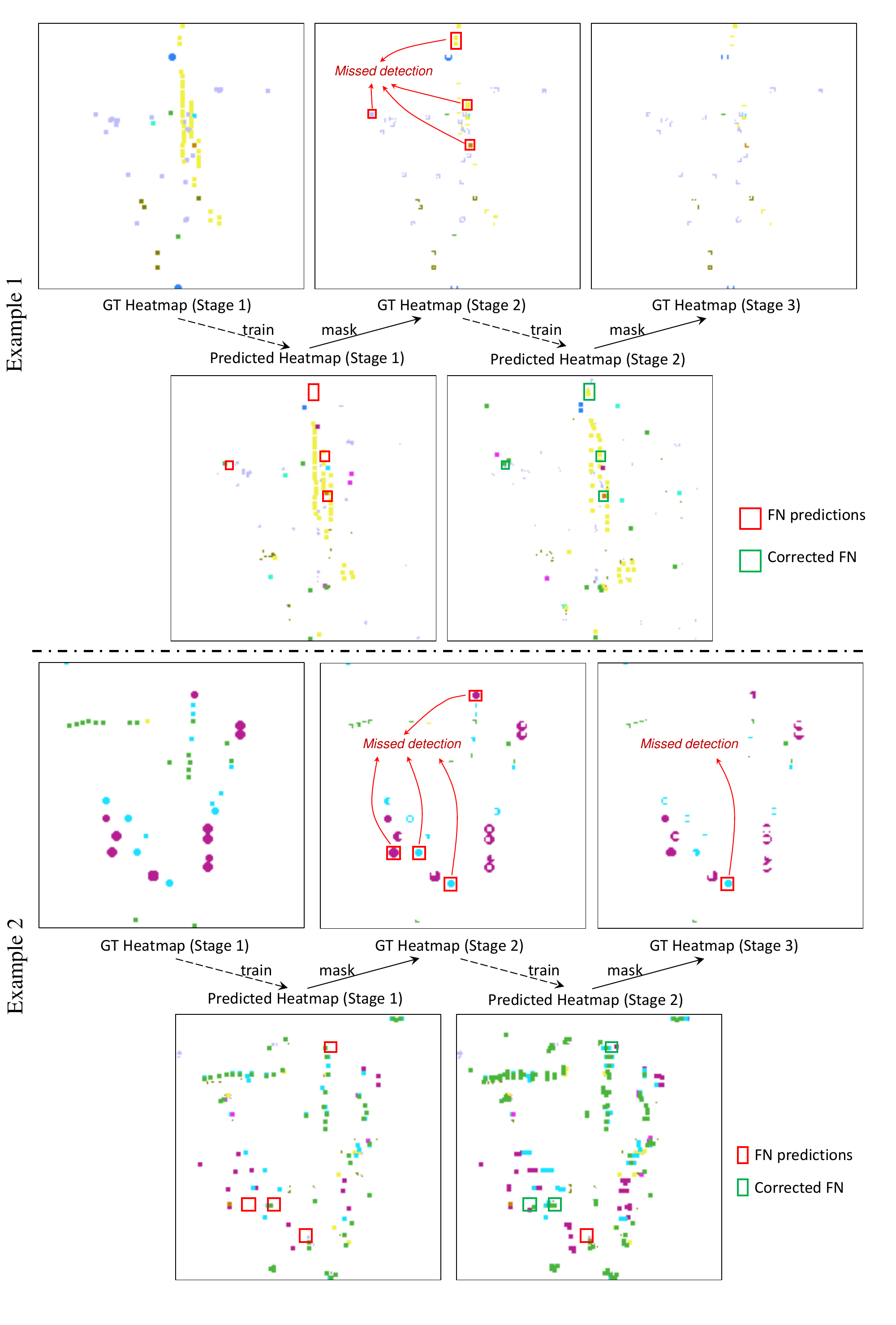}
	\end{center}
	\caption{\textbf{Example visualization of multi-stage heatmap encoder process on the bird's eye view}. The process of identifying false negatives operates stage by stage. We show different categories with different colors for visualization. The top three subfigures display the ground-truth center heatmaps at each stage, highlighting the missed object detections. The two subfigures below display the positive mask that shows positive object predictions. The scene ids are ''4de831d46edf46d084ac2cecf682b11a'' and ''825a9083e9fc466ca6fdb4bb75a95449'' from the nuScenes \textit{val} set. We recommend zooming in on the figure for best viewing.}
	\label{fig: visualization heatmaps and masking}
\end{figure*}

\vspace{2mm}
\noindent\textbf{Example visualization of multi-stage heatmaps and masking.}
We present a visual illustration of the multi-stage heatmap encoder process in Fig.~\ref{fig: visualization heatmaps and masking}. 

\begin{figure*}[t]
	\begin{center}
		\includegraphics[width=0.85\linewidth]{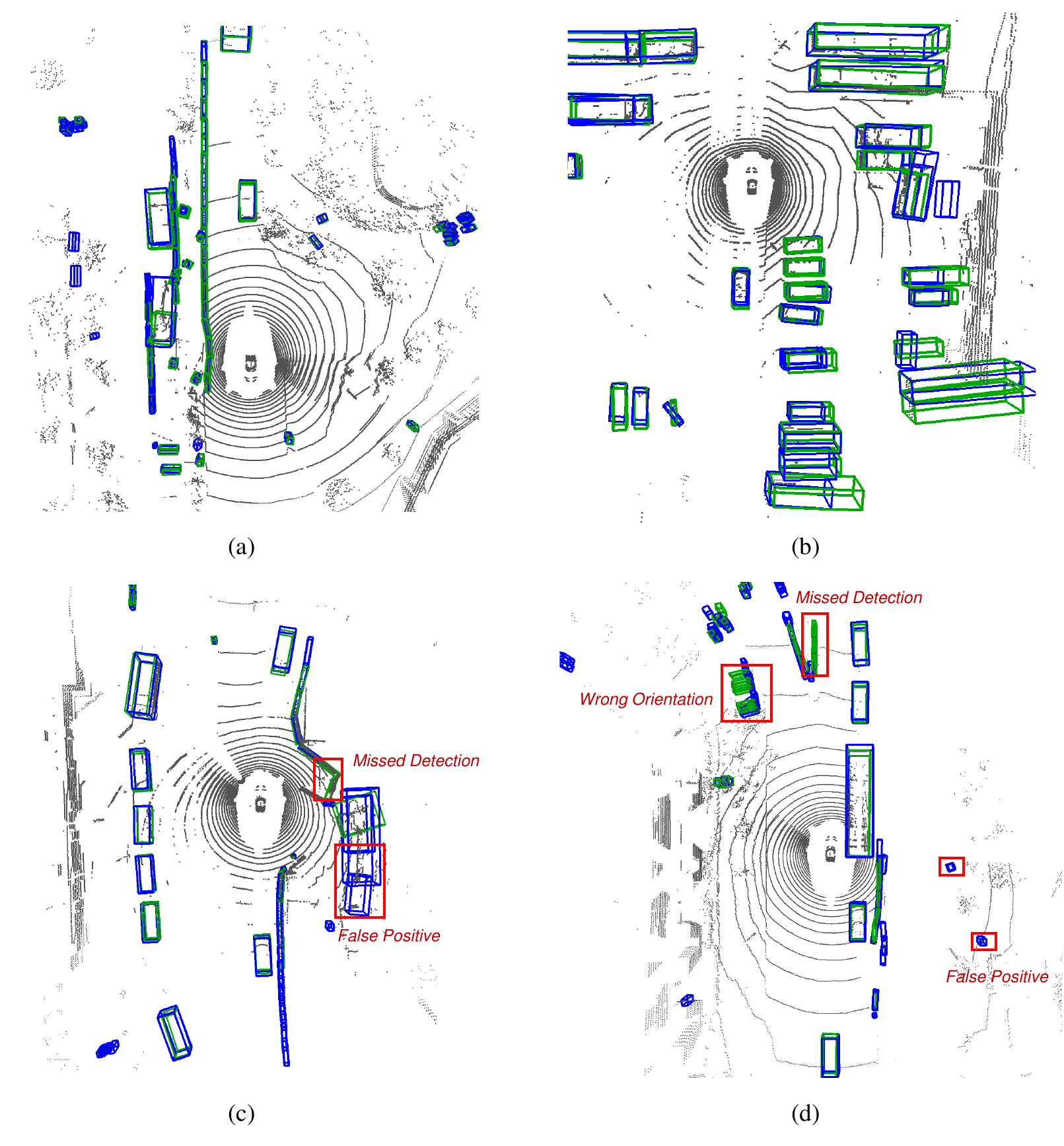}
	\end{center}
	\caption{\textbf{Visual results and failure cases.} The green boxes represent the ground truth objects and the blue ones stand for our predictions. We recommend zooming in on the figure for best viewing.}
	\label{fig: point cloud vis}
\end{figure*}

\vspace{2mm}
\noindent\textbf{Qualitative results.}
Fig.~\ref{fig: point cloud vis} shows some visual results and failure cases of FocalFormer3D on the bird's eye view. Although the average recall AR$_{<1.0m}$ reaches over $80$\%, some false negatives are still present due to either large occlusion or insufficient points. Also, despite accurate center prediction, false negatives can arise due to incorrect box orientation. Further exploration of a strong box refinement network is left for future work.

\end{document}